\documentclass[final,3p,times]{elsarticle}

\usepackage{amssymb}
\usepackage{url,hyperref,lineno,microtype}
\usepackage[onehalfspacing]{setspace}
\usepackage{arydshln}
\usepackage[gen]{eurosym}
\usepackage{booktabs}

\journal{Elsevier}

\begin{document}

\begin{frontmatter}



\title{A Machine Learning Approach for Player and Position Adjusted Expected Goals in Football (Soccer)}


\author[inst1]{James H. Hewitt}

\affiliation[inst1]{organization={Cardiff University, School of Computer Science and Informatics}, addressline={Abacws, Senghennydd Road}, city={Cardiff}, postcode={CF24 4AG}, country={UK}.}

\author[inst1]{Oktay Karaku\c{s}}

\begin{abstract}
Football is a very result-driven industry, with goals being rarer than in most sports, so having further parameters to judge the performance of teams and individuals is key. Expected Goals (xG) allow further insight than just a scoreline. To tackle the need for further analysis in football, this paper uses machine learning applications that are developed and applied to Football Event data. From the concept, a Binary Classification problem is created whereby a probabilistic valuation is outputted using Logistic Regression and Gradient Boosting based approaches. The model successfully predicts xGs probability values for football players based on 15,575 shots. The proposed solution utilises StatsBomb as the data provider and an industry benchmark to tune the models in the right direction. The proposed ML solution for xG is further used to tackle the age-old cliche of: 'the ball has fallen to the wrong guy there'. The development of the model is used to adjust and gain more realistic values of expected goals than the general models show. To achieve this, this paper tackles Positional Adjusted xG, splitting the training data into Forward, Midfield, and Defence with the aim of providing insight into player qualities based on their positional sub-group. Positional Adjusted xG successfully predicts and proves that more attacking players are better at accumulating xG. The highest value belonged to Forwards followed by Midfielders and Defenders. Finally, this study has further developments into Player Adjusted xG with the aim of proving that Messi is statistically at a higher efficiency level than the average footballer. This is achieved by using Messi subset samples to quantify his qualities in comparison to the average xG models finding that Messi xG performs 347 xG higher than the general model outcome. 
\end{abstract}




\end{frontmatter}


\section{Introduction}
Football is one of the lowest-scoring games, due to its single-scoring system where every goal is worth one point compared to for example Rugby where the minimum isolated event is worth three points (a penalty kick). This means that every chance is highly valuable. This emphasises the importance of being able to convert goals from scoring opportunities. Having predictions and probabilities of each chance has been proven to have a significant competitive advantage. If players' impact is only judged on converted goals when they are so rare, then, it is very possible to miss their actual impact. 

The main focus of this paper is to create and apply an Expected Goals (xG) model ‘from scratch’ and predict xG values with new and highly informative features. xG can be seen as a probability, ranging between 0 and 1, stating the chance that each shooting opportunity is converted into a goal. With the value always being within the 0-1 range but never 0 or 1, this measure suggests that a certain Goal (xG = 1.0) or certain No Goal (xG = 0.0) is not feasible. The model details include such events as the method in which the player receives the ball, the technique they use to shoot and then painting freeze frame pictures of all possible factors influencing the shot output quality. Shown below is a freeze frame example of a shot:
\begin{figure}[htbp]
    \centering
    \includegraphics[width=0.75\linewidth]{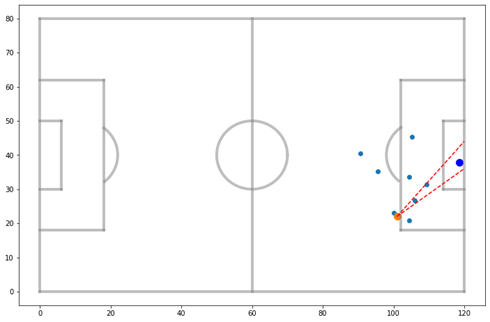}
    \caption{A freeze frame example: shot location (distance and angle) }
    \label{fig:freeze}
\end{figure}

Figure \ref{fig:freeze} details factors to be considered later on: shot location (distance and angle), opposition players surrounding the shooter, and opposition between the shot and goal. From this base level, this paper will document the development in the analysis of expected goals (xG). Finalising with novel \textit{Position} and \textbf{Player adjusted xG}, with the aim of providing industry competitive advantages and improvement in academic knowledge through this publication.

\section{Academic Look to the xG Metric}
The expected goals metric is quite new with a large gap in published papers. This is due to the football industry keeping technological developments in-house to gain a competitive advantage against opponents. Up-to-date data is also privatised and requires large funding to acquire and test which is increasing the barriers to producing academic papers within the xG subject area. 

One of the seminal pieces that started a certain data revolution is ``The Expected Goals Philosophy" by James Tippett\cite{tippett2019expected}. It discusses the development of the xG over time and highlights the evidence supporting xG’s application to football within performance and recruitment (scouting). Tippett \cite{tippett2019expected} begins discussing the ``dialect" and analysis of football as ``deeply flawed" and states that other sports such as Baseball, Basketball, Cricket and Hockey have already embraced strong mathematical approaches to performance analysis. 

Players' xG values are covered and analysed to show the xG totals for individual players against their total goals. A player with a high xG score suggests that they are often in goal-scoring positions, with this compared to total goals to judge how efficient they are with chances. This allows the reader to see if they are clinical, which would show a player over-performing their xG, if they are wasteful, or if they were underperforming their xG. Examples show Sadio Mane (Former Liverpool, now Bayern Munich) overperforming by an ``incredible" 5.24 xG in the 2018/19 Premier League season. The application of the xG values helped clubs such as Brentford and Brighton recruit ``undervalued players" and sell on for a profit. For example, Neil Maupay was transferred from St Etienne to Brentford prior to the 2017/18 season for a fee of \euro{}2 Million. Brentford then signed off on a further transfer for Maupay only two years later just before the 2019/20 season which saw Maupay move to Brighton for a fee of \euro{}15.56 million (TransferMarkt). This transfer fee of 678\% inflated from the original value Brentford paid, along with Maupay being pivotal and scoring 41 goals in 95 appearances (TransferMarkt) for Brentford during their stay in the English Championship. The Maupay funds were invested into Ivan Toney in September 2020, who incidentally led the side to back-to-back playoff finals. From a transfer fee of \euro{}5.6 million, he helped promote the side and now is sitting on a \euro{}45 million market value (TransferMarkt) after two successful seasons in the premier league. 

At the time of writing, Tippett suggested improvements worth considering in xG models, saying that the then OptaSport model did not incorporate defender positioning into the xG output. Also, xG models do not include threatening attacks that do not result in a shot.

Herold \emph{et. al.} published a summary paper that provides interesting insight into machine learning applications of football \cite{herold2019machine}. The authors discuss the lack of context to xG models built at that time. ``They either did not acknowledge or capture opponent positioning and thus, failed to provide context that coaches and analysts can apply to the match.”

Lucey \emph{et. al.} \cite{lucey2015quality} state that shots and shots on target values do not provide the true value of shot attempts using Spatio-temporal patterns to analyse the 10-second build-up to shot attempts. This paper assigns defender proximity-to-shooter to develop validity and to evaluate the shot quality and therefore determine whether a team was ``dominant" or ``lucky" with the outcome of the game. Further questioning whether the context of the scenario is important and proposing how probabilities change depending on the player attacking with regards to the opponent they are competing against within a contextual period of xG creation. 

Brechot and Flepp \cite{brechot2020dealing} deal with randomness in match outcomes with a large sample size of 7304 matches across four seasons in the top 5 leagues. Variables utilised include distance, angle, rule setting and body part. The empirical model used is a logistic regression for a binary response variable, finding that xG is the optimal source of a team’s actual performance based on xG scorelines in comparison to the number of points collected in those games. Allowing objective assessment of performance and underperformance, Brechot and Flepp \cite{brechot2020dealing} state that short-run events are often based upon randomness and therefore unsustainable. Allowing clubs be able to judge their players properly based on statistical performance rather than result outcome, therefore avoiding adjustments which may inhibit long-run performance and therefore results. The paper suggests accepting and working with randomness in football and using expected goals to navigate around its impacts, to allow sturdy decision-making processes. 

In a university thesis from Universitat Politècnica de Catalunya (UPC) in accordance with FC Barcelona by Madrero \cite{madrero2020creating}, the authors utilise location-based, contextual data such as previous pass method and detailing the area of the pass whether inside or outside the box, and player related information based on the shot body part along with some further player-related information scraped from FIFA games. They propose using three models: Logistic Regression, XGBoost and Neural Networks. 
Distance to goal proves to be the most influential feature, regardless of player quality, but better ‘rated’ players have a larger shot rating that they can accumulate higher xG values and goals. From the La Liga sample, Messi is shown as the highest scorer and the highest in terms of cumulative xG. In the same sample, there is also team-based xG, where the number of goals and accumulated xG correlate positively with the teams' position in the table. 

In Fairchild \emph{et. al.} \cite{fairchild2018spatial}, based on a sample of 1115 non-penalty shots from 99 games in the Major League Soccer (MLS - USA Football League), a logistic regression model is utilised with extracted features of shot location coordinates, distance from the goal line, the angle from vertical posts, shot type (categorical variable - \emph{cat.}), assist type (\emph{cat.}) and play type (\emph{cat.}). This paper concludes that spatio-temporal movements of players are essential along with fractal dimensions of ball trajectory. 
In Cavus and Biecek \cite{cavus2022explainable}, based on a large sample of 315,430 shots over seven seasons for European Top 5 leagues, the authors use an ‘explainable’ artificial intelligence approach to create ‘explainable expected goals’ with the aim of producing accurate expected goals for the team and player performance analysis. The features used are game minute, home or away fixture, play type (situation e.g. Open Play, Set Play), shot type (limb), last action (e.g. cross), distance from goal and angle from goal. They use tree-based classification: XGBoost, Random Forest, Light GBM and CatBoost, and also propose utilising Aggregated Profiles (AP) to demonstrate the difference in model predictions depending on a change in the value of a feature.

Umami \emph{et. al.} \cite{umami2021implementing} use factors such as distance and angle of shot on goal. Utilising Distance and Angle as one combined variable, this paper demonstrates that it had a greater impact on calculating the xG. They also state that the process could be applied to further research to allow understanding of distance and angles influence on gaining and giving up xG and in turn goals. 
Another interesting development they suggest is an implementation of more difficult quantified variables such as surface type and condition, role play, and confidence.


Tureen and Olthof \cite{tureen2022estimated}, with a sample of 580 Premier League and 326 Women’s Super League fixtures, use Statsbomb data provided for the 2022 conference and aim to quantify individual players' nested data in a hierarchal structure to reduce biased interferences. By creating the Estimated Player Impact (EPI) measure using the Generalized Linear Mixed Models (GLMM), they estimate individual players' impact on each xG valuation and are able to quantify the shot impact on shot conversion. The EPI metric quantifies a player’s impact on xG estimations. Similarly, to this paper, the EPI is measured on positional subgroups: Forward, Midfielder and Defenders (split into central and wing backs). Their findings claim that Heung Min Son of Tottenham Hotspurs who plays as a forward has the largest positive impact on EPI per xG.

\section{XG in Football Industry}
The analytics revolution and widespread usage of xG are something relatively new. Statistics were somewhat shunned by classical old-fashioned pundits in the original stages. Possession stats are being heavily used to indicate ‘running of a game’ despite some teams being set up to counterattack and therefore choosing to have less possession of the ball. This possession-orientated school of thought was quickly dismissed. However, in 2017 the then Arsenal Manager Arsene Wenger, following a defeat to Manchester City stated:
“If you look at the expected goals, it was 0.7 for them and 0.6 for us, it was a very tight game, they created very little, had a very little number of shots on target, 0.1 more than us, that’s all.” \cite{winterburn_2017}. This was not greeted well as Arsenal had lost that game 3-1. But what was going to happen from there was xG would only build traction. 

At this time, it was absolutely correct for viewers and fans to be sceptical about the validity of xG in football. At this point in time, Sky Sports ran a report detailing early stage xG and specifically how “Expected goals also struggles to account for where the defenders are on the pitch - something that can dramatically impact the relative chances of two shots taken from precisely the same spot” \cite{bate_campbell_2017}. Over the next few years, xG became more prominent with companies such as StatsBomb and Opta heavily utilising the metric in their data packages.

Underlying advanced metrics and analytics such as xG provide utility for Performance Analysis teams within the professional game, Recruitment Departments within those teams and also Bettors/ Betting Companies. Performance analysts refer to objective information used by athletes, coaches and analyst departments. Performance analysis includes Video analysis of the game tape, including coaching analysis of tactics. Then, data-driven decisions are based on in-game performance such as pass completion rates, and sports science-related criteria such as heart rate monitor statistics. Analysts gain from xG by applying it to their own team and opposition analysis in preparation or games using event and track data to build a number-driven approach to the game. xG allows the performance department objective evaluation on player performance, moving away from ‘he should’ve scored that’ to ‘statistically that's a huge chance’. 
Another application is gaining a competitive advantage by building visualisations that graphically show the areas in which xG is given up by opposing defences. This allows the potential implementation and exploitation of weaknesses in the opposition's defence. Goals may seem quite random to a casual view, however, xG can graphically provide a basis to utilise angles, distances and densities of a shooting zone and how that influences a potential goal. The idea of implementation of xG allows a different perspective for performance analysts, prior to the data revolution of the 2010s, was limited to game tape analysis and biased views on the difficulty of chances. 

Effective recruitment analysis provides possibly the largest competitive advantage in all sports. A player's ability to be in the right zones and shooting positions is more indicative of potential success. This is where xG comes in. Finding high cumulative xG footballers at a young age allows cheaper signing and room for development. Brentford's rise is heavily linked to xG and this approach is detailed in \cite{tippett2019expected}. 
Being able to identify undervalued players creates room for profits and improved team quality on a low budget. For attackers and attacking midfielders, xG as an underlying metric of quality can provide insight into which players may be worth signing prior to a goal-scoring run.

Betting companies also need the usage of xG, if a team is not conceding a lot but receiving a lot of xG against then its models will pick up on the trend. It would be fair to presume that a large element of luck behind their lack of goals was conceded considering a lot of xG against them. Likewise, if a team is not conceding goals and not conceding xG then it is safe to presume that the team is effectively defending. These underlying statistics allow a Betting Company to hedge its profit margins against values that in the past were not available to individual consumers. 

As discussed in \cite{tippett2019expected}, SmartOdds and Matthew Benham used xG to make a stamp on the football industry. They first used expected goals, prior to its mainstream implementation, to ‘win millions’ each year by placing bets on football fixtures. With this same money, Benham purchased shares in Brentford and then used xG as a recruitment parameter and signed, as discussed before, Neil Maupay who funded a positive chain of events for the West London side. SmartOdds and Benham are living proof of the success of the application of xG.

\section{The Proposed xG Model Details and Results}

Based on the academic and football industry background of xG discussed above, this paper aims to 
\begin{enumerate}
    \item to create a reliable and robust xG model on open play goals. 

    \item to evaluate models based on their abilities to predict goals close to actual goals. This means that the direction of the project is not to build the best model according to classifier performance. But compare classifier outputs with real-life events. 

    \item to develop a model the output of which is useful in showing the quality of players' finishing to a high degree.

    \item to achieve a significant positive correlation with industry xG providers.

    \item to successfully apply adjusted xG metrics to specific matches.
\end{enumerate}

For the purpose of reaching the aforementioned objectives, in this section, we share the proposed xG model along with the developed data set. We then share our results under four experimental scenarios of (1) a general model comparison of the baseline and proposed model, (2) position-adjusted xG analysis, (3) player-corrected xG analysis and (4) an industry benchmark testing based on the 2018 Champions League Final game of Liverpool versus Real Madrid. 

\subsection{Data}
The whole data set in this paper has been web-scraped by using \textit{statsbombpy} Python module that Statsbomb provides. Please visit the Statsbomb GitHub page via \url{https://github.com/statsbomb} for details. The data frame is filtered to include just open-play shots, with penalties removed as they skew the model. Leaving a sample of 15,575 data points of event data. The shot breakdown is 2873 European Championship and World Cup samples, 12,688 La Liga samples with 2,301 being Messi shots and then a Game Sample of 28 shots from the Champions League. A separate xG model can be run to calculate penalty xG, however, this is not a point of interest in this project. The industry standard of 0.766 xG per penalty is an acceptable value. 

From the event data sample discussed, there are 95 original features whereby 8 were selected to be used in the model along with all the open-play shots. The features were reduced by removing all variables that had over 99\% NaN or Null values. From there, features were assessed and selected if they were viewed to directly impact a shot. For the supervised learning of a binary classification problem, some steps are taken to prepare the data for learning. Any missing values are replaced with 0, numeric features are scaled, categorical features are label encoded, Boolean features are one hot encoded and coordinates are separated into X and Y. In total, after pre-processing, there are 26 variables in this initial model with 1 target variable of Goals (1 = Goal, 0= No Goal).


Based on these variables selected, an xG model is built to provide a baseline for development within the paper with the idea of surpassing the Baseline model by including additional variables and further model tuning.

\subsection{xG Model}
A logistic regression approach was taken to establish a baseline xG model. The aim is to create a model that shows the minimum acceptable level and then improve the results from there. Logistic regression is a traditional approach as it is easy to run, interpret and train. 
With 
the sample of 15,575 open-play shots, the baseline model is tested versus the industry benchmark of StatsBomb xG. Then, Accumulated Goals are used as a metric to compare to Accumulated xG with the aim that the correlation would be high, to judge whether the model created is close to the industry level of expected goals, and that accumulated Goals and xG are close. 

\begin{figure}[ht!]
    \centering
    \includegraphics[width=0.49\linewidth]{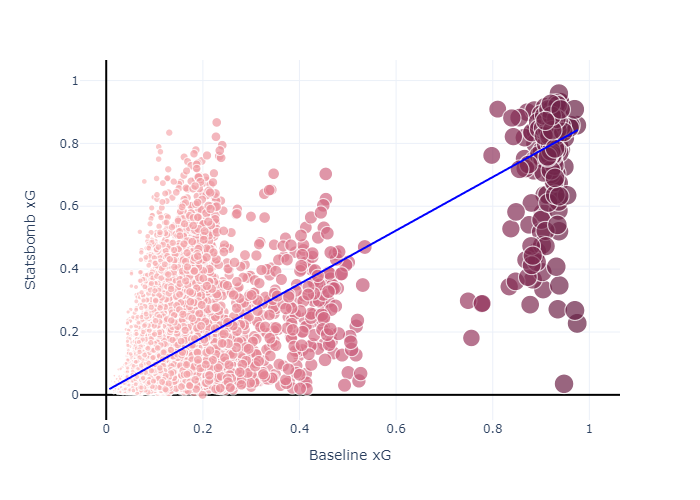}
    \includegraphics[width = 0.49\linewidth]{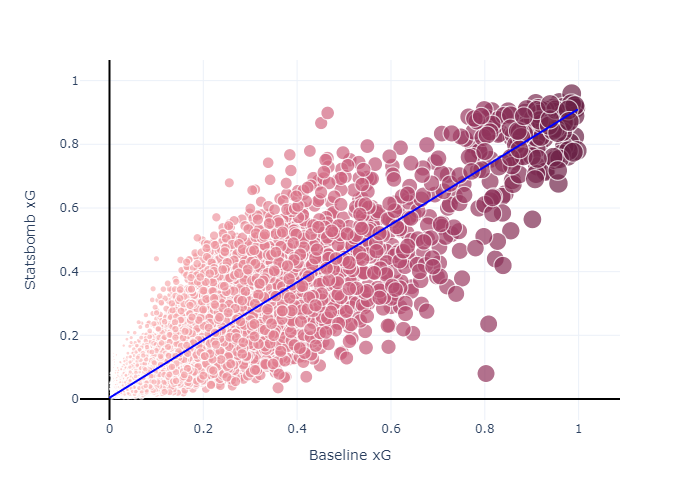}
    \includegraphics[width = 0.49\linewidth]{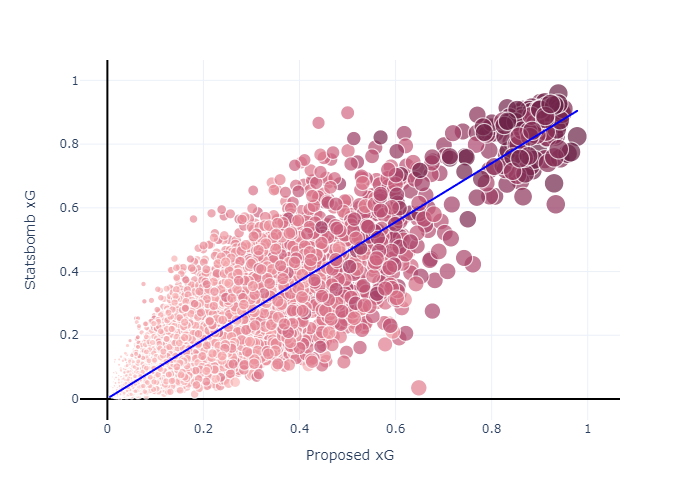}
    \caption{(a) Baseline model results with existing 26 features. (b) Baseline model performance after adding extra features. (c) Proposed xG model performance.}
    \label{fig:baseline}
\end{figure}

Figure \ref{fig:baseline}-(a) represents the output of the baseline logistic regression xG model compared to the xG of StatsBomb. The fitted line is
$y = 0.01 + 0.85x$, and the correlation is 0.659. 
Graphically, there is evidence that the model is not working at maximum efficiency as there is a large area without any considerable xG values, with 15,575 shots it would be expected to see a more even distribution of shot xG values and not so skewed to the two tails. In conclusion, the distribution of xG values suggests that it is limited in predicting a certain level of expected goals. The graph suggests that the xG Model is under-predicting xG values, hence the large cluster on the left-hand side between 0.0 xG and 0.2 xG.

The lower performance of the existing baseline is not caused by the model capability but by the limited amount of information provided to the model with the existing features. From the event data sample discussed, there were originally 95 features whereby 8 were selected to be used in the model along with all the open-play shots. The features were reduced by removing all variables that had over 99\% NaN or Null values which resulted in a total of 26 features. Some of those are \textit{Location, Shot-technique, Pass-type} and \textit{Direction}. It is clear from the analysis in Figure \ref{fig:baseline}-(a) that existing features are not enough to provide a robust xG model. 

Parallel with the suggestions from the experts such as \cite{tippett2019expected}, we developed various new features to better inform the proposed xG model. Some examples of additional features utilised are Goalkeeper positioning, Player Pressure Radiuses and Opposition Between the shot and the goal. Figure \ref{fig:baseline}-(b) presents the same baseline modelling performance but this time with the new features along with the existing ones (40 features in total). 
Please see Table \ref{tab:feat} for the 8 Statsbomb data variables and ``some" of the proposed engineered variables for the purposes of this paper. Following the creation and implementation of further variables, the logistic regression results are extremely promising and run extremely quickly. With the benchmark being StatsBomb xG, the line outcome is $y = 0.00 + 0.90x$ with a correlation of 0.887. Visually, the distribution is also pleasing and the output is acceptable. The predicted total xG is at 1866 compared to the actual goals of 1887. 

\begin{table}[htbp]
  \centering
  \caption{Predictors and their data types.}
    \begin{tabular}{p{3cm}p{2.75cm}|p{5.5cm}p{3cm}}
    \toprule
    \multicolumn{2}{p{5.75cm}|}{\textbf{StatsBomb Variables}} & \multicolumn{2}{p{8.5cm}}{\textbf{Engineered Variables}} \\
    \midrule
    Aerial shot & \textit{Binary} & Strong Footed & \textit{Categorical} \\
    1st time & \textit{Binary} & Within Penalty Area & \textit{Binary} \\
    Open Goal & \textit{Binary} & Distance from Goal & \textit{Scaled Distance} \\
    Pressure & \textit{Binary} & Angle to centre of goal & \textit{Scaled Angle} \\
    Shot Body Part & \textit{Categorical} & GK distance from centre of goal & \textit{Scaled Distance} \\
    Shot Technique & \textit{Categorical} & Opp. players between shot and goal & \textit{Value} \\
    Pass Type & \textit{Categorical} &       &       \\
    Location & \textit{Coordinates} &       &        \\
    Goal & \textit{Binary \textbf{(Target)}}  &       &       \\
    \bottomrule
    \end{tabular}%
  \label{tab:feat}%
\end{table}%

The purpose of this paper is to extend and improve the baseline approach with a decision trees-based deep learning approach. Decision trees are a non-parametric supervised learning method used for classification and regression. Decision tree algorithms are prominent in current Kaggle competitions and various regression-based prediction studies. With the goal of creating a model that predicts the value of a target variable by learning simple decision rules inferred from the data features, the supervised learning gradient boosting family models are promoted. 

Gradient boosting aims to combine weaker models (performing only slightly better than random choice), building upon Leslie Valiant’s  probably approximately correct (PAC) learning \cite{valiant1984theory}, which investigated the complexity of machine learning problems. Gradient tree boosting is the statistical framework, where the aim is to minimise the total loss of the models. This is achieved by combining weaker learning models using gradient descent. It has three main areas of focus (1) Loss function, (2) Weak learner and (3) Additive model. 

Figure \ref{fig:baseline}-(c) depicts the regression analysis plot for \textit{the decision-trees based proposed xG model}. Examining Figure \ref{fig:baseline}-(c), we can conclude that the proposed xG model is slightly better than the logistic regression. The visualisation shows extremely promising results with the fitted line of $y = 0.00 + 0.92x$ and a correlation of 0.902 with the StatsBomb xG. Predicting an accumulated xG of 1870 out of the 1887 actual goals as shown in Table \ref{tab:goals}. 
\begin{table}[ht]
    \centering
\caption{Model comparison based on the number of goals predictions.}
    \begin{tabular}{p{2cm}p{3cm}p{3cm}p{3cm}p{3cm}}
    \toprule
         &  Baseline & Proposed & Statsbomb & Exact Goals\\
         \toprule
       Goals  & 1866 & 1870 & 1751 & 1887\\
       \bottomrule
    \end{tabular}
    \label{tab:goals}
\end{table}

\subsection{Results}
The proposed xG model was tested on three different
perspectives :
\begin{enumerate}
    \item We first presented position-adjusted xG analysis.
    \item Subsequently, a similar analysis to (1) but this time player-adjusted xG analysis was implemented using Lionel Messi as the target player.
    \item Finally, we performed an industry benchmark on a specific fixture of R. Madrid vs. Liverpool in the 2018 UEFA Champions League Final. 
\end{enumerate}

\subsubsection{Position Adjusted xG}
For the purposes of this experimental step, we start by analysing the data set by separating the players in terms of their stronger positions, such as Goalkeeper, Defender, Midfielder, and Forward. The positional analysis results are presented in Table \ref{tab:pos1}.

\begin{table}[htbp]
    \centering
 \caption{Positional analysis of the data set.}
    \begin{tabular}{p{2cm}p{2cm}p{2cm}p{2cm}p{2cm}p{2cm}p{2cm}}
    \toprule
Position    &	Total Shots & Total Goals&	Shot/Goal& Baseline xG& Statsbomb xG&	Proposed xG \\\toprule
Forward &	8646&	1276&	6.77&	1265.767&	1154.624&	1252.802\\
Midfield&	4590&	398&	11.53&	399.3236&	391.2747&	411.0013\\
Defender&	2336&	213&	10.97&	200.9173&	205.6996&	206.8463\\
Goalkeeper	&2	&0	&0	&0.096735		&0.107875&0.105983\\\bottomrule 
    \end{tabular}
    \label{tab:pos1}
\end{table}

The results show some interesting outcomes. Forwards are known as the most clinical finishers in football, hence why they are entrusted with the position closest to the opposition goal with the most chances to score. Their conversion rate supports this, at 6.77 shots per goal. As expected goals represent the average probability of the shot turning to a goal, you should predict that the forwards are finishing at a rate higher than the average. Forwards scored 1276 Goals in this data, with all reference xG predictions being lower than the actual goals, suggesting an over-performance. 

Midfielders show something quite different compared to Forwards. In all the models, they are underperforming by 1 and up to 13 below their xGs. This suggests that midfielders are wasteful with their chances. From the optimal selected model, midfielders are performing at 411 xG and scoring 398 goals. Defenders' results are surprising even based on the average xG including some high-value attempts by talented forwards. All xG predictions for defenders are below the exact number of goals, which suggests a high level of clinical nature and an overperformance of xG.  

\begin{figure}[t!]
    \centering
\includegraphics[width=0.49\linewidth]{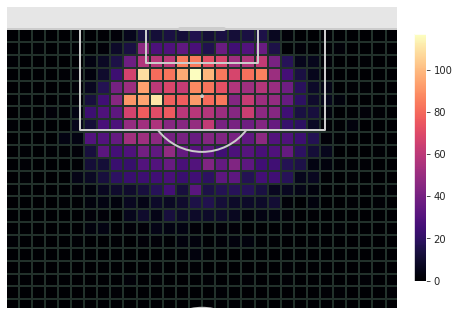}
\includegraphics[width=0.49\linewidth]{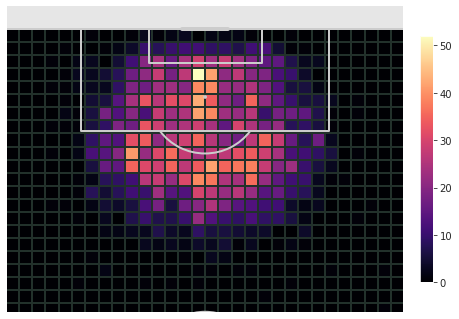}
    \includegraphics[width=0.49\linewidth]{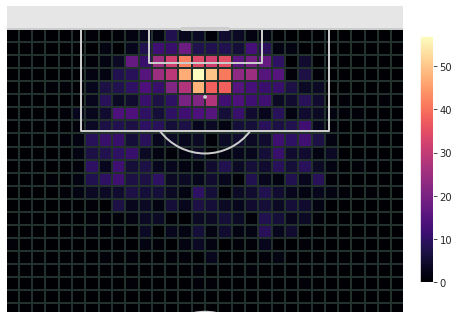}
    \caption{Shot Density of (a) Forwards, (b) Midfielders, (c) Defenders.}
    \label{fig:positionHeat}
\end{figure}

Figure \ref{fig:positionHeat} depicts the density of shots for each position except GKs. From Figure \ref{fig:positionHeat}-(a), as expected, the density of shots by forwards is predominantly within the box which suggests higher xG opportunities and therefore potentially more goals. Figure \ref{fig:positionHeat}-(b) shows a very different story compared to the forwards with a lot of low density being in and around the six-yard box, which is where forwards showed the highest density. The shots from outside the box are at a high density and suggest that shooting from the range is more prominent with midfielders compared to any other subgroup. Long shots do not have high xG chances and may suggest why their conversion rates from xG to actual goals are lower than Forwards. From Figure \ref{fig:positionHeat}-(c), the density of the shots is as expected that the defenders' main chances come from set pieces and corners with the highest level of density falling between the penalty spot and the six-yard box. Considering this, the higher conversion rate and goals surpassing xG are very interesting to see in comparison to Midfielders, who football fans would consider more skilled in converting goals. 

As discussed in this section up to now, the results gathered suggest a deviation in xG performance between positional subgroups. To more accurately quantify the performance of each, the same classification problem is carried out. However, in this instance, the training dataset is changed on three different occasions. The model is now trained on each positional sub-group, creating Forward Adjusted xG, Midfield Adjusted xG and Defender Adjusted xG. The aim is to provide empirical evidence that each position has different levels of efficiency over a whole scale of data. 

\begin{table}[htbp]
    \centering\caption{Positional Adjusted xG Values}
    \begin{tabular}{p{3cm}p{2.2cm}p{2.2cm}p{2.2cm}p{2.2cm}p{2.2cm}}
    \toprule
Model	& xG & Forward xG&	Midfield xG&	Defender xG&	Goals\\
\toprule
&&&&&\\
Goals/xG	&1870	&1956	&1728	&1397	&1887\\
&&&&&\\\hdashline
&&&&&\\
Adjustment Value	&0	&+86	&-142	&-473	&0\\
&&&&&\\\bottomrule

    \end{tabular}
    \label{tab:pos2}
\end{table}

Table \ref{tab:pos2} shows, as can be easily expected by an average football fan, Forward Adjusted xG predicts the highest value, followed by Midfield and then Defenders.  Forwards xG valuation increases by an absolute adjustment value of 86, this large increase suggests that if all the chances are changed to being taken by a forward-skilled player then xG increases by 86. For Midfielders, when their sample skillset is applied across the whole dataset they find a reduction from the average xG of 142. This result is an expected outcome, as most clinical players (presumed to be forwards) are removed from their training set. Defenders, follow a similar yet more drastic reduction in xG even in comparison to Midfielders. Their xG reduces by 473 and therefore shows that the model predicts a lower clinical ability and conversion rates to other positional groups. This is what was expected by the testing, and proves that the model is working effectively.

\subsubsection{Player Adjusted xG Values}
Along with the aforementioned positional adjusted xG, this paper also aimed to test player-specific xG. The motivation of player-adjusted xG is to assess how much better elite-level forwards are at finishing in comparison to normal players and average xG. 

\begin{figure}[t]
    \centering
    \includegraphics[width = 0.8\linewidth]{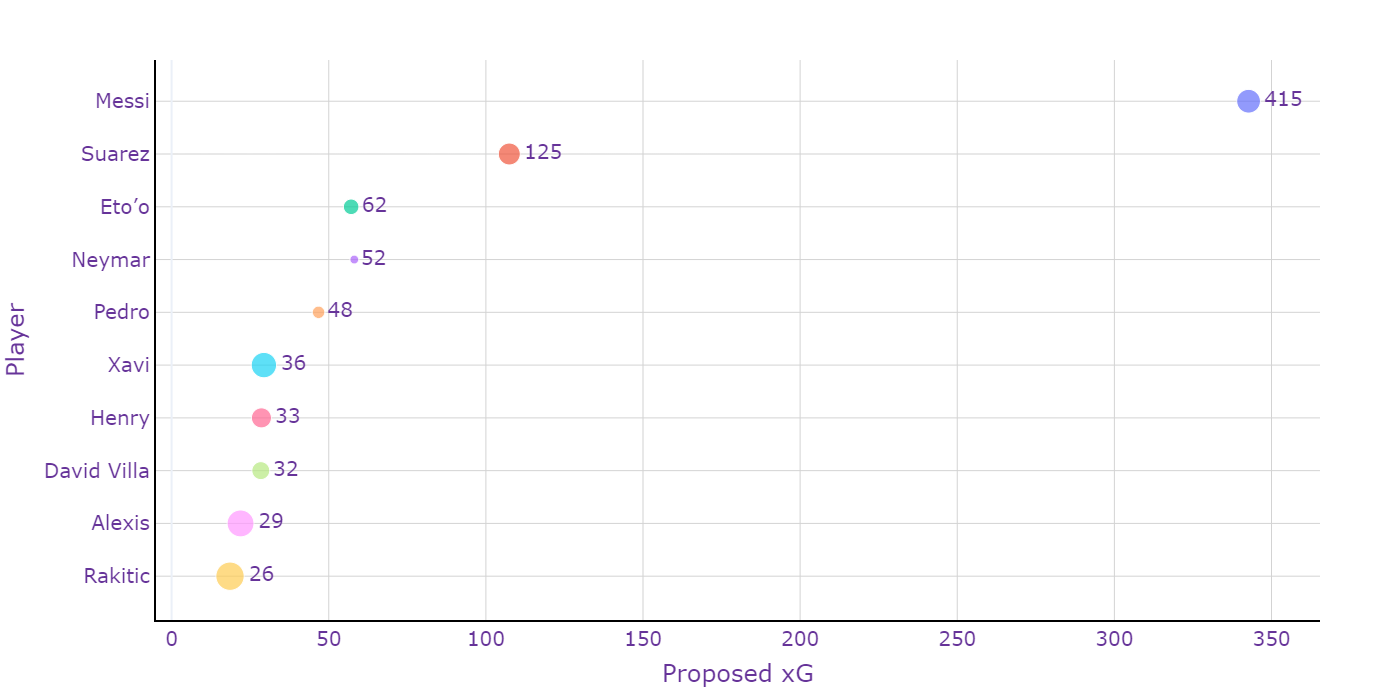}
    \caption{Player specific xG and Goals analysis. The size of the markers shows Goals per xG metric whilst the text next to each marker is the exact number of goals scored by the player.}
    \label{fig:player}
\end{figure}

The utilised data set in this paper shows that most elite finishers produce Goals greater than their xG value. Lionel Messi's sample of 415 Goals from 342 xG (\%21.34 higher) is a large overperformance whilst Luis Suarez has 125 goals with 107 xG (\%16.82 higher). Messi's overperformance is to a degree larger than any other player in the sample at 0.82 xG per Goal (Please see Figure \ref{fig:player} for more details). 

Considering the data set in this paper includes a large sample of Lionel Messi fixtures in La Liga and therefore a lot of Messi shots and therefore goals, we decided to use Lionel Messi as the player adjustment test case. Thus, we use Messi's 2300 shots and 415 goals to train all xG models. Another reason why Messi is selected as he is widely known as one of the all-time greats and has significantly overperformed xG in all xG models. He is a great sample to compare the ``average" player against and allows us to show how significant his skill set is.

The process was similar to the positionally adjusted xG in that a subset of the data is taken with it being made up of only Messi shots, this also leaves the other subset as a sample including No-Messi shots. From these two subsets, Messi data is used to train an individual xG model, and No-Messi data is tested to evaluate the player-adjusted xG values.

The initial results of this analysis show significantly inflated levels of xG based on the Messi Adjusted xG. 
The results presented in Table \ref{tab:pla1} show that the Messi-trained model produces significantly more accumulated xG, at 1874 xG predicted which is an adjustment of 347 xG over the whole sample. This is as expected and confirms the hypothesis that Messi is an elite shooter. 

\begin{table}[htbp]
    \centering \caption{Messi Adjusted xG Values}
    \begin{tabular}{p{3cm}p{3cm}p{3cm}|p{5cm}}
    \toprule
Model	& Goals & Proposed xG&	Messi adjusted xG\\
\toprule
&&&\\
Score 	&1472	&1527	&1874\\
&&&\\\hdashline
&&&\\
Adjustment Value	&0	&0	&+347\\
&&&\\
\bottomrule
    \end{tabular}
    \label{tab:pla1}
\end{table}

Table \ref{tab:pla2} offers further insight into Messi’s influence on the models. Specific player-adjusted values show a dramatic improvement in all players' xG. This also means e.g. if Messi shot the shoots Luis Suarez had, he could accumulate 126.21 xG with a nearly 20 xG increase compared to the proposed model's initial result of total of 107.45 xGs. 
Significantly, based on the original model, every player in the sample, except Neymar, over-performed their xG suggesting that they are clinical and finish chances greater than the average. With Messi adjusted xG, now every player falls short of their xG values, suggesting that they are underperforming. 

\begin{table}[htbp]\centering\caption{Player-specific results for Messi Adjusted xG}
\begin{tabular}{p{2cm}p{2cm}p{2cm}p{2cm}|p{3cm}|p{3cm}}
\toprule
Player	& Goals & Proposed xG&	Statsbomb xG&	Messi adjusted xG & Over-performance?\\
\toprule
Suarez      & 125   & 107.4513  & 108.2466       & 126.2198 &No                          \\
Eto’o       & 62    & 57.1248   & 52.71371       & 63.02088    &No                      \\
Neymar      & 52    & 58.12527  & 56.02049       & 66.74318 &No                        \\
Pedro       & 48    & 46.75322  & 43.47725       & 53.60634  &No                       \\
Xavi        & 36    & 29.35869  & 28.75191       & 41.78057    &No                      \\
Henry       & 33    & 28.5776   & 28.18925       & 32.03867   &\textbf{Yes}                       \\
David Villa & 32    & 28.3766   & 29.0367        & 32.61256   &No                        \\
Alexis      & 29    & 21.9309   & 21.33671       & 25.03286    &\textbf{Yes}                       \\
Rakitic     & 26    & 18.57919  & 18.45674       & 26.59705   &No                     \\\bottomrule
\end{tabular}
    \label{tab:pla2}
\end{table}

\begin{figure}[ht!]
    \centering
    \includegraphics[width = \linewidth]{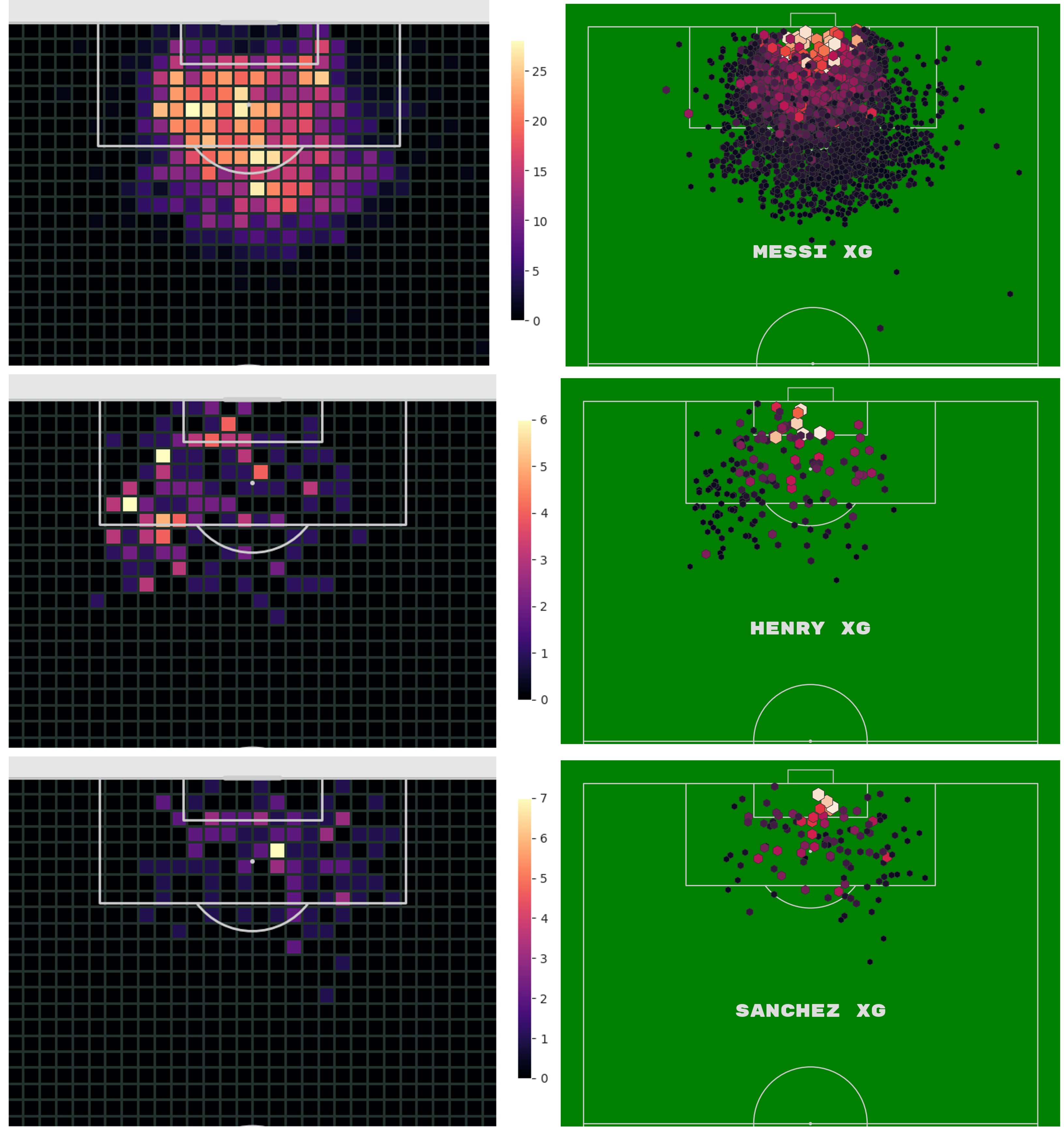}
    \caption{(LEFT) Shot location heatmap and (RIGHT) xG location map visualisations for Lionel Messi, Thierry Henry and Alexis Sanchez from top to bottom, respectively.}
    \label{fig:player2}
\end{figure}

Figure \ref{fig:player2} depicts shot location heatmaps and xG location maps for three important finishers namely Lionel Messi, Thierry Henry and Alexis Sanchez. In addition, Henry and Sanchez are the only two strikers who over-performed in Messi-adjusted xG values. All subfigures show how each player has some different shooting patterns whilst their xG maps are logical and aligned with the expected outcomes: The closer to the goal, The higher the xG. The left-hand side shot heatmap subfigures highlight how clinical Lionel Messi is with his capability to score over and outside of the box whilst Henry's trademark finesse shot to the far post can be seen with a high probability of shots from the left. It looks like Alexis Sanchez mostly tries shots from inside the box which gives him an over-performance on Messi-adjusted xG values. 

This trial emphasises the value one of the all-time greatest - Lionel Messi - adds to the accumulated probabilities and confirms the expectation that he is an Elite Level finisher and far above the average player in the sample, despite other high-quality finishers being in the sample such as Thierry Henry.
 

\subsubsection{Industry benchmark testing and application}
This experimental analysis details the proposed xG model being applied to the Real Madrid vs Liverpool, Champions League final from the 2017/18 season with the aim of proving the ability to apply the xG model to a specific fixture.

\begin{table}[htbp]
    \centering \caption{Game statistics for the 2018 Real Madrid vs Liverpool, Champions League final}
    \begin{tabular}{p{2cm}p{1.5cm}p{1.5cm}p{2cm}p{2cm}p{2cm}|p{2cm}}
    \toprule
         Team& Goals& Shots& Statsbomb& FBRef& infogol& \textbf{Proposed}  \\\hline
         &&&&&&\\
         Liverpool& 1& 14& 1.31442&	\textbf{1.9}& \textbf{1.88}& 1.61353 \\
                           &&&&&&\\
         R. Madrid& \textbf{3}& 14& \textbf{1.367858}&	1.5& 1.71& \textbf{1.816377}\\
         \bottomrule
    \end{tabular}
   
    \label{tab:clFinal1}
\end{table}

This game was finalised within 90 minutes with shots drawn at 14 apiece, R. Madrid beat Liverpool 3 goals to 1. In Table \ref{tab:clFinal1}, there are xG figures from StatsBomb (use their own data), FB Ref (use Opta data) and infogol.com along with the proposed xG model. Comparing and contrasting these figures allows the validity of the proposed model to be shown. Accumulated xG from StatsBomb is at 2.7, FBRef falls at 3.4, infogol is at 3.59 and the proposed model is at 3.4. This also shows the different factors within the models and feature tuning between companies. The results from the proposed model are still satisfactory as they are within the same region as all the companies, despite resource constraint differences. 

\begin{figure}[t!]
    \centering
    \includegraphics[width=0.75\linewidth]{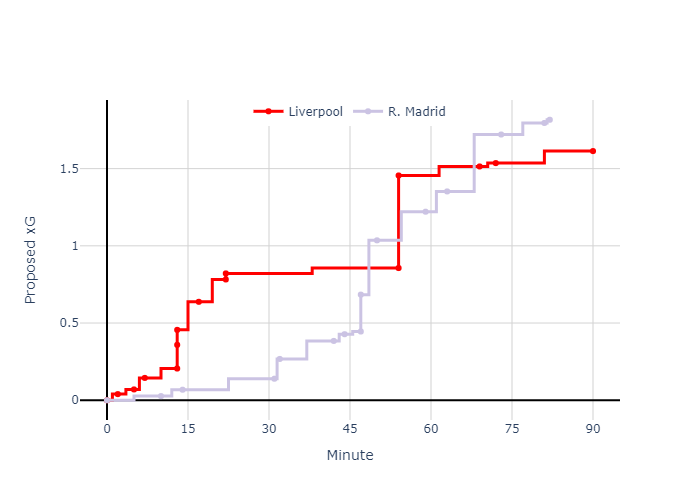}
    \includegraphics[width=0.65\linewidth]{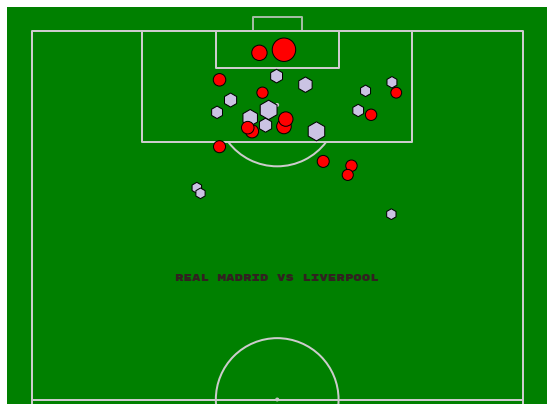}
    \caption{2018 Real Madrid vs Liverpool, Champions League final (TOP) xG timeline, (BOTTOM) xG location map.}
    \label{fig:clFinal}
\end{figure}

Figure \ref{fig:clFinal}-(a) shows an industry-standard recreation of the play-by-play xG events, known as an xG timeline, something commonly used to show a better understanding of which periods of the game were more productive for each side. Figure \ref{fig:clFinal}-(b) visualises each shot within this fixture using industry-standard recreation a shot map with the larger icon indicates a higher xG.
\begin{table}[htbp]
    \centering \caption{2018 Real Madrid vs Liverpool, Champions League final game goals.}
    \begin{tabular}{p{2cm}p{3cm}p{3cm}p{2cm}p{2cm}p{2cm}}
\toprule
Team&Player&Shot Technique&Minute&	Proposed xG& Statsbomb 
xG\\
\toprule
R. Madrid& Karim Benzema& Volley& 50	&0.351569	&0.517137\\
Liverpool&	Sadio Mane&	Volley&	54&	0.599426&	0.548516\\
R. Madrid&	Gareth Bale&	Overhead Kick&	63&	0.131235&	0.022605\\
R. Madrid&	Gareth Bale&	Normal&	82&	0.020848&	0.013965\\\bottomrule
    \end{tabular}
   
    \label{tab:clFinal2}
\end{table}

Table \ref{tab:clFinal2} details each goal within the game showing the xG valuation differences between the proposed and the StatsBomb xG models. The 2nd goal of the game by Sadio Mane and the 4th Goal of the game by Gareth Bale show similar valuations. However, Karim Benzema’s goal from the Karius mistake is slightly lower from the proposed model which is surprising as the accumulated xG showed that the proposed xG model quantified higher than StatsBomb. That trend however is shown in the Gareth Bale Overhead Kick, the proposed xG shows a 0.131 value whereas StatsBomb is at 0.022. In this scenario, the result is concerning as it likely suggests an over-prediction of the xG value of overhead kicks, which are known to be a very tough skill. 

\section{Conclusions}
In conclusion, the project achieves various important insights in terms of position and player adjusting of xGs with a machine learning model created from scratch along with various important new features. 
The results from Position Adjusted xG and Messi Adjusted xG aligns completely with the expectations and therefore are inferred to have predicted xG accurately. 

The main contributions to current literature are development in xG modelling and more features than are used in conventional models. Some of the additional features which are utilised in this paper are Goalkeeper positioning, Player Pressure Radiuses and Opposition Between the shot and the goal. These features directly develop what many writers including \cite{tippett2019expected} have specifically called for, to develop xG. The successful implementation and application of these allow further study such as creating a variable that can indicate the surface area that a player has available to shoot at which may require products such as StatsBomb 360 data. 

The key findings are that Forwards are the most clinical and most effective shooters from the positional subgroups, regardless of adjustment or not. Further developments find that the shooting quality order is Forwards, Midfielders, and Defenders in terms of their ability to successfully gain xG. The Player Adjusted xG finds that Messi performs better than even the Forward subset. 
This level of analysis on player and positional adjusted xG cannot be found in current literature, and we believe it successfully fills a large gap.

Further research could include adjusting xG for Leagues and European competitions using the same subset process that this project follows. Questions can be asked such as: ‘How much better are Premier League players in comparison to the top 5 leagues?’ and ‘How much more efficient are players performing in the Champions League subgroups to their Domestic Leagues?’. 

As further the opening paragraph, football is situation based and has room to adjust metrics to increase the validity of statistics within the professional game and improve performance analysis.

\bibliographystyle{elsarticle-num} 
\bibliography{main}





\end{document}